\documentclass{article}

\PassOptionsToPackage{compress}{natbib}
\usepackage{graphicx}
\usepackage{listings}
\usepackage[preprint]{neurips_2024}

\usepackage[utf8]{inputenc} 
\usepackage[T1]{fontenc}    
\usepackage{hyperref}       
\usepackage{url}            
\usepackage{booktabs}       
\usepackage{amsfonts}       
\usepackage{nicefrac}       
\usepackage{microtype}      
\usepackage{xcolor}         

\title{LightAgent: Production-level Open-source Agentic AI Framework}

\author{\\%
  Weige Cai\textsuperscript{1},
  Tong Zhu\textsuperscript{2},
  Jinyi Niu\textsuperscript{3},
  Ruiqi Hu\textsuperscript{1},
  Lingyao Li\textsuperscript{1},
  Tenglong Wang\textsuperscript{4},\\
  Xin Guo\textsuperscript{1},
  Zhaowei Liu\textsuperscript{1},
  Dongpo Cheng\textsuperscript{1},
  Xiaowu Dai\textsuperscript{2},
  Weining Shen\textsuperscript{2},
  Liwen Zhang\textsuperscript{1*}\\
  \\
   \textsuperscript{1}Shanghai University of Finance and Economics\\
\textsuperscript{2}University of California Los Angeles\\
\textsuperscript{3}Fudan University\\
\textsuperscript{4}Shanghai University\\
}

\begin{document}
\def\thefootnote{*}\footnotetext{Corresponding authors.}\def\thefootnote{\arabic{footnote}}

\maketitle

\begin{abstract}
 With the rapid advancement of large language models (LLMs), Multi-agent Systems (MAS) have achieved significant progress in various application scenarios. However, substantial challenges remain in designing versatile, robust, and efficient platforms for agent deployment. To address these limitations, we propose \textbf{LightAgent}, a lightweight yet powerful agentic framework, effectively resolving the trade-off between flexibility and simplicity found in existing frameworks. LightAgent integrates core functionalities such as Memory (mem0), Tools, and Tree of Thought (ToT), while maintaining an extremely lightweight structure. As a fully open-source solution, it seamlessly integrates with mainstream chat platforms, enabling developers to easily build self-learning agents.  We have released LightAgent at
\href{https://github.com/wxai-space/LightAgent}{https://github.com/wxai-space/LightAgent} .
\end{abstract}

\section{Introduction}
\setlength{\parindent}{2em} 

\indent Multi-agent systems represent a paradigm shift in artificial intelligence by enabling collaborative problem solving through coordinated interactions among multiple specialized agents (Wang et al.,
2023; Xi et al., 2023). The rapid advancement of Large Language Models (LLMs)  (Ouyang et al., 2022; Touvron et al., 2023a,b) has catalyzed the development of multi-agent applications across domains such as software engineering (Hong et al., 2023), societal simulation (Park et al., 2023), and intelligent assistants (Wu et al., 2023). However, despite these advancements, critical challenges persist in designing versatile, robust, and efficient platforms for multi-agent system development. These include managing agent specialization, standardizing dynamic workflows, ensuring fault tolerance against LLM errors (Rawte et al., 2023), and integrating multi-modal data and tools systematically (Lewis et al., 2020a; Yao et al., 2023). Existing frameworks often struggle to balance flexibility with simplicity, hindering rapid deployment and scalability.

To address these challenges, we introduce LightAgent, an extremely lightweight yet powerful framework designed to streamline multi-agent application development. LightAgent redefines efficiency by combining minimalist architecture with advanced capabilities, including autonomous tool generation, multi-agent collaboration, and robust fault tolerance. Unlike conventional platforms burdened by dependencies like LangChain or LlamaIndex, LightAgent achieves full functionality in a 100\% Python implementation with a core codebase of only 1,000 lines, ensuring rapid deployment across diverse scenarios.

The framework addresses three critical dimensions of multi-agent systems: versatility, robustness, and scalability. First, LightAgent supports dynamic agent specialization through customizable memory modules (mem0), enabling personalized long-term user interaction tracking and autonomous learning. Second, its built-in Tree of Thought (ToT) module enhances complex task decomposition and reflection, while the LightSwarm subsystem simplifies intent-driven multi-agent collaboration, surpassing traditional swarm-like approaches in flexibility. Third, LightAgent pioneers automated tool generation: by ingesting API documentation, developers can generate hundreds of domain-specific tools within one hour, drastically reducing development cycles.

Notably, LightAgent advances robustness through LLM-powered error detection and self-correction mechanisms, mitigating risks of hallucination (Zhang et al., 2023b) and tool execution failures. The framework natively supports multi-modal data handling, tool unification, and knowledge-sharing strategies, while its compatibility with major LLMs (e.g., OpenAI, ChatGLM, Qwen) and streaming API integration ensures seamless adoption in real-world applications. By harmonizing lightweight design with systematic innovation, LightAgent establishes a new benchmark for accessible, scalable, and resilient multi-agent systems.

This paper details LightAgent's architecture, demonstrating its efficacy in reducing development complexity while enhancing agent autonomy. We further validate its performance through case studies highlighting rapid tool generation, cross-agent collaboration, and adaptive learning capabilities. Our contributions pave the way for democratizing advanced multi-agent applications across research and industry.

\section{Related Work}
The evolution of multi-agent systems (MAS) has been significantly shaped by advancements in distributed artificial intelligence and complex system modeling. This section reviews key contributions in chronological order to highlight the progression of MAS research.

\subsection{Early Foundations and Architectural Frameworks}
The conceptual foundation of MAS was established through early works on agent autonomy and interaction protocols. Jennings (2001) introduced fundamental principles of agent-based software engineering, emphasizing autonomy and social ability as core characteristics of intelligent agents (Jennings, 2001). This was extended by Wooldridge (2009) through a comprehensive taxonomy of MAS architectures, identifying three primary orientations: information-flow, control-oriented, and role-oriented designs  (Wooldridge, 2009).

Significant architectural contributions emerged in 2013 when Tianfield et al. proposed an autonomic architecture for complex MAS, introducing novel coordination mechanisms for self-organizing systems (Tianfield et al., 2013). Their work laid groundwork for subsequent industrial applications by addressing system decomposition and component interdependencies.

\subsection{Application-Specific Developments (2015-2020)}
Between 2015-2020, MAS found increasing adoption in domain-specific implementations:

\begin{itemize}
    \item \textbf{Recommender Systems}: Morais et al. (2019) demonstrated MAS's effectiveness in personalized recommendation engines, employing agent-based collaborative filtering for dynamic user modeling (Morais et al., 2019).
    
    \item \textbf{Industrial Automation}: Leitão et al. (2017) pioneered the integration of MAS with cyber-physical systems in manufacturing, achieving distributed production scheduling through agent negotiation protocols (Leitão et al., 2017).
    
    \item \textbf{Energy Management}: Pinto et al. (2019) developed a CBR-based MAS for smart grid optimization, showcasing 18\% efficiency gains in energy distribution through multi-agent coordination  (Pinto et al., 2019).
\end{itemize}

\subsection{LLM-Driven Architectures (2021-Present)}
Recent breakthroughs in large language models (LLMs) have revolutionized MAS design:

\begin{itemize}
    \item \textbf{MetaGPT} (2024): This SOP-driven framework introduced role-specific agents (e.g., product managers, developers) with structured communication protocols, achieving 73\% task completion accuracy in software development simulations  (Anonymous, 2024).
    
    \item \textbf{AutoGen} (2024): Microsoft's conversational agent framework enabled dynamic group decision-making through its unique \texttt{ConversableAgent} class, demonstrating 40\% faster consensus-building in creative tasks compared to single-agent systems  (Research, 2024).
    
    \item \textbf{XAgent} (2025): Featuring a dual-loop architecture with PlanAgent and ToolAgent components, this system achieved state-of-the-art performance in complex task decomposition (89\% success rate) through iterative plan optimization  (Anonymous, 2025).
\end{itemize}

Current research directions emphasize hybrid architectures combining LLM capabilities with traditional MAS principles, as seen in recent industrial applications of swarm intelligence  (Group, 2024) and adaptive manufacturing systems (Latsou \& Farsi, 2021).

\section{Overview}

\subsection{Basic Concepts in LightAgent}

\subsubsection{Agent}

In artificial intelligence, an intelligent agent is an entity equipped with task-specific capabilities powered by LLMs, that perceives its environment, takes actions autonomously to achieve goals, and may improve its performance through machine learning or by acquiring knowledge. Each agent operates independently but collaborates with other agents as needed.

\subsubsection{Memory}

In agent models, memory holds a highly significant role as the core component. Memory records and stores users' historical interaction data, enabling agents to review past conversations during multi-turn dialogues. This capability allows agents to continuously self-optimize, avoid repeating mistakes, and deliver personalized responses tailored to users' specific needs and usage habits. Furthermore, in the context of complex task management, memory can conveniently document the implementation process at each stage, including user feedback and operational records. This not only enhances task execution efficiency but also aids agents in reusing experiences and refining strategies when facing similar issues in the future.

LightAgent features support for the external memory module(mem0), which automates context retention and historical record management without manual intervention. This memory mechanism allows agents to store and retrieve user-specific information, ensuring context consistency across conversations.

In LightAgent, long-term memory is continuously updated with each interaction with the user, and agents automatically manage the memory content. This ensures the long-term validity and colnsistency of the memory. By providing custom long-term memory for each user, LightAgent enhances conversation consistency and personalization, significantly improving the user experience, and reducing the need for manual intervention in memory management.

\subsubsection{Tools and Tool Generator}

Tools are predefined or dynamically generated utilities that agents can use to enhance their problem-solving abilities. Agent with tools can overcome the limitations of its own model by leveraging external or pre-configured resources to perform various tasks. Additionally, the object-oriented design enhances the clarity and modularity of the system architecture, benefiting both developers and users. Our framework supports both tool importing and automated tool generation to ensure versatility across tasks.

\subsubsection{Multi-Agent Collaboration}
Multi-agent systems are formed by the organic integration of multiple agents into a cohesive whole. In contrast to a mere aggregation of agents, in a multi-agent system the individual agents frequently exchange and share information. Each agent can independently complete tasks while also being capable of collaborating when necessary. By decomposing tasks and working collaboratively, a multi-agent system has the ability to solve more complex problems, and through effective division of labor, it achieves complementary advantages that lead to more efficient problem-solving.

Within our framework, LightAgent emphasizes coordination between agents, enabling efficient task delegation and collaborative problem-solving. Through the exchange of information and cooperative work among agents, LightAgent is able to break down complex tasks into multiple sub-tasks that are processed in parallel by different agents, significantly increasing both the speed and accuracy of problem resolution. Additionally, this collaborative mechanism encourages agents to discuss new issues and share experiences, continuously optimizing their decision-making strategies and avoiding the limitations and repetitive mistakes that may occur with a single agent.

\subsection{Framework}
In our framework, we create a LightSwarm based on the task at hand, and automatically register several LightAgents. Then, synchronize the relevant information with memory. Next, we parse the intent of the task to plan a Tree of Thought for answering the question, and flexibly decide whether agent collaboration is required. Once the Tree of Thought is established, we begin generating the answer and appropriately invoke relevant tools to produce the output. For more complex tasks that require multi-turn dialogue, we store the previous conversation information and further iterate to arrive at the final output.
\begin{figure}[ht]
    \centering
    \includegraphics[width=\textwidth]{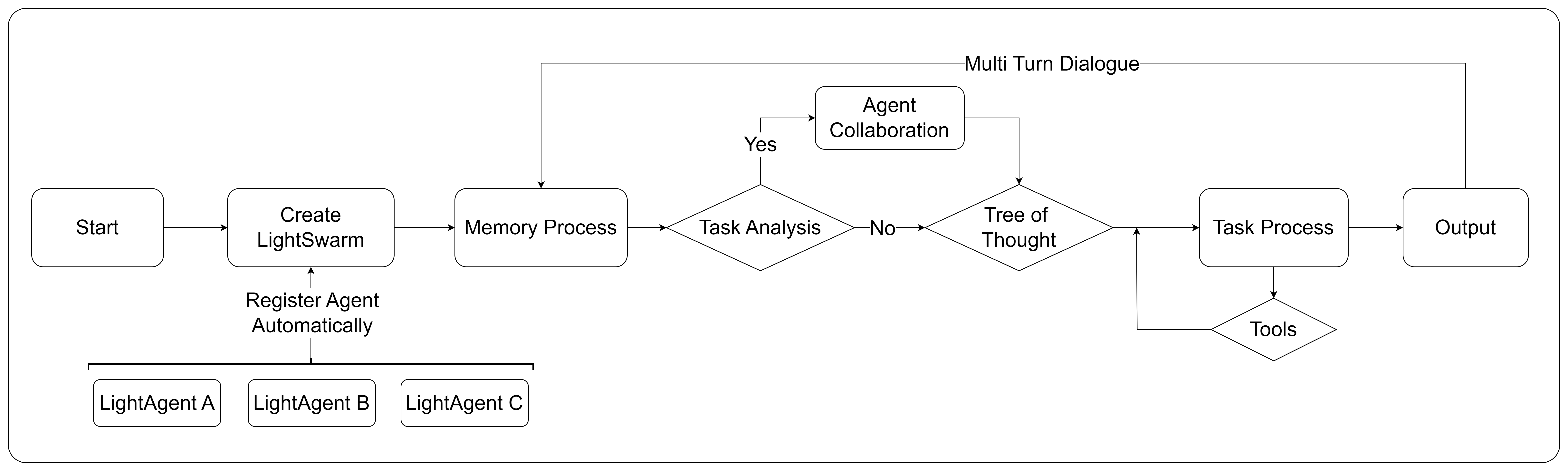}
    \caption{Framework of LightAgent}
    \label{fig:figureframework}
\end{figure}

\section{Lightweight and Adaptable}

LightAgent is designed to be lightweight, adaptable, and highly efficient. Unlike existing frameworks, it eliminates the need for complex external libraries such as LangChain and LlamaIndex. The framework is implemented entirely in Python, LightAgents ensures broad compatibility with a wide range of large language models (LLMs), including OpenAI (Achiam et al., 2023), Zhipu ChatGLM (TeamGLM, 2024), DeepSeek (Liu et al., 2024), Qwen series (Bai et al., 2024) large models. Additionally, LightAgent supports OpenAI streaming format API service output, enabling integration with mainstream chat frameworks.

LightAgent adopts a minimalist design philosophy, resulting a highly lightweight and intelligent system. It is built with a focus on architectural simplicity, reduced code complexity, and minimal external dependencies. These design principles make it adaptable and efficient for variety of use cases. Its key advantages are highlighted in the following aspects:

\subsection{Low resource consumption}
LightAgent is a framework implemented 100\% Python-based framework with a core codebase of just 1,000 line. It avoids reliance on complex third-party libraries, making it suitable for embedded devices, low-power environments, and applications with strict resource requirements.

Despite its powerful functionality, LightAgent maintains low latency and high processing efficiency during operation, making it suitable for real-time data processing and interactive applications.

\subsection{Easy Deployment}
LightAgent can be quickly deployed on almost any device or environment without a cumbersome configuration process, ensuring quick and efficient startup. With minimal dependencies and a straightforward configuration process, it can be set up in minutes. 
\begin{lstlisting}[caption=Installation]
# Install the latest version of LightAgent
pip install lightagent
# (Optional installation) Install the Mem0 package via pip:
pip install mem0ai

# Hello World Example Code
from LightAgent import LightAgent

# Initialize Agent
agent = LightAgent(model="gpt-4o-mini", api_key="your_api_key", base_url="your_base_url")

# Run Agent
response = agent.run("Hello, who are you?")
print(response)
\end{lstlisting}
An active developer community ready to assist and provide answers at any time. Whether through detailed documentation, troubleshooting guides, or real-time support on forums, users can quickly resolve issues and optimize their deployments. This strong community backing ensures that LightAgent remains accessible to developers. We also release our code and welcome contributions from the community. The instructions for contributing are available in the GitHub repository.

\section{Configurable Agent}

\subsection{Agent Customization}
LightAgent provides a powerful agent customization feature that enable users to define each agent’s behavior, role, and toolset to meet specific requirement. This flexibility allows agents to be tailored for a wide range of tasks, from data analysis and information retrieval to API integration and beyond. By configuring different tools, users can create agents that are highly specialized and optimized for their unique use cases.
\begin{lstlisting}[caption=Customized Agent]
import requests
from LightAgent import LightAgent

def search_news(
        keyword: str,
        max_results: int = 5
) -> str:
    """
    Search news based on keywords
    """
    results = f"By searching for {keyword}, I've found {max_results} related pieces of information."
    return str(results)

# Custom Tools
tools = [search_news]  # including tool

# Initialize Agent
# Replace with your model parameters, API key, and base URL
agent = LightAgent(model="qwen-turbo-2024-11-01", api_key="your_api_key", base_url="your_base_url", tools=tools)

\end{lstlisting}

\subsection{Memory and Learning Transfer}

Agents record historical task experiences through the mem0 memory module. These experiences are transferred and applied to new tasks, enabling the system to make more efficient decisions when facing new challenges.

\begin{lstlisting}[caption=Customized Memory]
# Enable Memory Module

from mem0 import Memory
from LightAgent import LightAgent
import os
from loguru import logger

class CustomMemory:
    def __init__(self):
        self.memories = []
        os.environ["OPENAI_API_KEY"] = "your_api_key"
        os.environ["OPENAI_API_BASE"] = "your_base_url"
        # Initialize Mem0
        config = {
            "version": "v1.1"
        }
        
        self.m = Memory.from_config(config_dict=config)

    def store(self, data: str, user_id):
        """Store memory."""
        result = self.m.add(data, user_id=user_id)
        return result

    def retrieve(self, query: str, user_id):
        """Retrieve related memory."""
        result = self.m.search(query, user_id=user_id)
        return result

agent = LightAgent(
        role="Please remember that you are LightAgent, a useful assistant to help users use multiple tools.",  # system role description
        model="deepseek-chat",  # Supported models: openai, chatglm, deepseek, qwen, etc.
        api_key="your_api_key",  # Replace with your large model provider API Key
        base_url="your_base_url",  # Replace with your large model provider api url
        memory=CustomMemory(),  # Enable memory function
        tree_of_thought=False,  # Enable Chain of Thought
    )

# Memory-enabled test & if tools need to be added, you can add tools to the agent for memory-enabled tool calls
\end{lstlisting}

\subsection{Autonomous Learning for Collaboration}
Autonomous learning is one of the core capabilities of ahents in LightAgent, enabling it to continuously accumulate experience, evaluate feedback, and optimize task execution strategies without external intervention. This process primarily involves the system’s ability to improve its learning strategies gradually through multiple interactions, thus enhancing its performance in long-term tasks.
Through autonomous learning, LightAgent not only adapts to the changing needs of users but also enhances its long-term task execution capabilities and intelligence through continuous reflection and adjustment.

\begin{lstlisting}[caption=Agent Initialization and Query Execution Code]
agent = LightAgent(
        name="Agent A",  # Agent name
        instructions="You are a helpful agent.",  # Role description
        role="Please remember that you are LightAgent, a useful assistant to help users use multiple tools.",  # system role description
        model="gpt-4o-mini",  # Supported models: openai, chatglm, deepseek, qwen, etc. qwen-turbo-2024-11-01 \ step-1-flash
        api_key="your_api_key",  # Replace with your API Key
        base_url="http://your_base_url/v1",  # API URL
        memory=CustomMemory(),  # Enable memory function
        self_learning=True,  # Enable agent self-learning
        debug=True,
        log_level="DEBUG",
        log_file="example.log"
    )

user_id = "test_user_1"
query = "I now have a procurement payment that needs to be transferred. What is my approval process?"
agent.run(query, stream=False, user_id=user_id)
query = "Please remember: According to the new company regulations, starting from January 2025, all procurement payments must first be signed by Manager Ding, who is responsible for procurement, then submitted to the finance manager for approval. After the finance manager's approval, the general manager of the company must also approve before the cashier can make the payment."
agent.run(query, stream=False, user_id=user_id)

user_id = "test_user_2"
query = "Hello, I have a procurement payment to transfer to the other party. How do I apply for the transfer?"
agent.run(query, stream=False, user_id=user_id)

\end{lstlisting}

\subsection{Automated Tool Generation}
Tools in LightAgent are designed to improve problem solving by serving as predefined utilities or dynamically generated resources. Its automated tool generation feature lets users quickly create tools that meet their needs, eliminating the need for redundant manual coding. This ability to automatically generate tools enables LightAgent to quickly adapt to various needs, greatly enhance development efficiency, and minimize human errors.

A key advantage of LightAgent is its ability to generate tools automatically based on user-provided API documentation or descriptions. This dramatically reduces development time and effort, allowing users to focus on higher-value tasks rather than building tools from scratch. The auto-generated tools are highly versatile, supporting both simple functionality extensions and seamless integration into existing agent systems. This adaptability ensures LightAgent can handle a wide range of tasks while meeting unique and evolving requirements.

By integrating these tools into LightAgent, users empower agents to dynamically select and utilize the most appropriate tools for specific tasks. This enhances task execution efficiency and ensures greater flexibility in handling complex demands.

\section{Multi-Agent Collaboration}
Multi-agent collaboration is one of the standout features of LightAgent, where multiple agents work in coordination to efficiently solve complex tasks. This multi-agent collaboration mechanism allows LightAgent to fully mobilize the capabilities of each agent in the system, improving overall task processing efficiency and effectiveness when handling large-scale and complex tasks.

Collaborative Work: Multiple agents can work together, leveraging their respective strengths and specialties to complete tasks. For example, some agents focus on data collection, while others are responsible for analysis, and the results are ultimately consolidated.

Intelligent Collaborative Decision Making: Through the LightSwarm module, LightAgent enables collaborative decision-making between agents. During cooperation, agents can share information, synchronize decisions, and avoid redundant operations and conflicts.

\subsection{Task-Oriented Agent Switching}
In a multi-agent collaborative system, tasks can be dynamically allocated and transferred based on the workload and capacity of the current agent, ensuring optimal allocation of resources.

\subsection{DeepSeek-R1 Based Agent Inference Planning ToT Engine}



The DeepSeek-R1 method relies on the following core steps, each of which provides systematic support for solving complex problems:

\begin{enumerate}
    \item \textbf{Problem Definition:} Clearly define the core problem and objectives. Initially, a thorough analysis of the problem is required to clarify the key issues and goals, ensuring that subsequent analysis and actions are directed appropriately.
    \item \textbf{Information Gathering:} Systematically collect relevant data and information. This process involves gathering not only all known data related to the problem but also extending to potential sources that may impact the solution.
    \item \textbf{Problem Decomposition:} Break down the complex problem into multiple sub-problems or modules. This step helps to decompose the large problem into smaller, more manageable parts, allowing for a more detailed analysis.
    \item \textbf{Multi-Dimensional Analysis:} Analyze each sub-problem from various angles and perspectives. This phase ensures that the problem is examined thoroughly from multiple viewpoints to avoid overlooking any critical factors.
    \item \textbf{Establishing Connections:} Identify and analyze the relationships and dependencies between sub-problems. This step reveals the inherent connections between issues, providing a foundation for integrating and optimizing solutions.
    \item \textbf{Solution Generation:} Propose potential solutions for each sub-problem. This process emphasizes creative thinking to generate multiple response strategies, offering a range of options for decision-making.
    \item \textbf{Evaluation and Selection:} Evaluate the feasibility, impact, and risks of all proposed solutions, selecting the most appropriate one. This phase focuses on assessing both the practical effectiveness and long-term consequences of each solution.
    \item \textbf{Implementation and Feedback:} Implement the selected solution and adjust based on feedback. This phase ensures that real-time feedback is utilized to adjust the solution, ensuring the achievement of the desired outcome.
\end{enumerate}

\begin{lstlisting}[caption=Structure of Tree of Thought ]
# Enable Tree of Thought
agent = LightAgent(
    model="qwen-turbo-2024-11-01", 
    api_key="your_api_key", 
    base_url="your_base_url", 
    tree_of_thought=True,  # Enable Tree of Thought
    tot_model="deepseek-r1", 
    tot_api_key="sk-uXx0H0B***17778F1",  # your deepseek r1 API Key
    tot_base_url="https://api.deepseek.com/v1",  # api url
)
\end{lstlisting}

The DeepSeek-R1 framework provides a clear, systematic path for solving complex problems. Through structured thinking and flexible strategies, the method significantly enhances the efficiency and accuracy of problem-solving.

\section{Future Work}
LightAgent is an extremely lightweight active agentic framework equipped with memory (mem0), tools (Tools), and a Tree of Thought (ToT) structure. With its focus on low resource consumption and high flexibility, LightAgent has successfully empowered developers to build scalable multi-agent systems.  The future development of LightAgent will focus on three directions to further enhance its capabilities in real-world applications: \\ (1) Adaptive Tool Mechanism; \\ (2) Memory-Enabled Agent Collaboration;\\ (3) Agent Assessment.
\subsection{Adaptive Tool Mechanism}
As illustrated in the LightAgent framework, multi-agent systems and a wide variety of tools are utilized to accomplish tasks. To address challenges across different disciplines and examine the same issue from multiple perspectives, it is essential to have access to a broad range of tools. However, an excess of tools can result in unnecessary token usage and computational power consumption. Therefore, our future plan is to allow users to add an unlimited number of tools, while enabling the large model to first select a candidate set of tools from thousands of tools. Then our model will filter out irrelevant tools before submitting the context to the large model.

\subsection{Memory-Enabled Agent Collaboration}
A multi-agent and the aggregation of single-agent models have significant differences in their mechanisms. The reason why a multi-agent model outperform single-agent models in terms of functionality is that a multi-agent system can communicate and share information with each other. This internal communication between agents enables more efficient collaboration, allowing them to collectively tackle complex tasks that a single agent would struggle to handle.

In multi-agent mechanism, the ability to share information and pass messages creates a dynamic and flexible environment, where agents can work together to solve problems that require different types of expertise or diverse approaches. This collaboration is particularly beneficial for tasks that involve coordination, negotiation, or the need for diverse perspectives. By enabling agents to coordinate their actions, adapt to changing environments, and leverage the knowledge and skills of other agents, multi-agent systems offer superior scalability, efficiency, and adaptability compared to single-agent models.

In the future, as we continue to enhance LightAgent’s capabilities, we aim to refine this collaborative agent mechanism, allowing for even more sophisticated communication and cooperation among agents. This will open up new possibilities for solving complex, multi-faceted problems across various fields, from autonomous vehicles and robotics to large-scale data analysis and decision-making.


\subsection{Agent Assessment}
An embedded agent evaluation tool is provided to facilitate the assessment and optimization of constructed agents. This tool will provide real-time feedback on key performance metrics such as response accuracy, efficiency, task success rates, and adaptability, ensuring that each agent aligns perfectly with specific business scenarios. Furthermore, this tool is designed to be user-friendly, seamlessly integrated with our model, and enables industry users to easily conduct self-assessments using our framework.

\section{Conclusion}
In this paper, we propose \textbf{LightAgent}, a lightweight and flexible framework specifically designed to simplify and enhance the deployment of multi-agent systems powered by large language models. By integrating core features—including memory (mem0), customizable tools, and Tree of Thought (ToT)—LightAgent successfully addresses the common trade-off between simplicity and versatility. Its fully open-source nature facilitates seamless integration with mainstream chat platforms, significantly lowering entry barriers for developers.

Future work will explore optimizing adaptive tool-selection mechanisms to reduce computational overhead, enhancing collaborative memory capabilities among agents, and refining the built-in assessment system to align closely with real-world business scenarios. Ultimately, we hope LightAgent fosters broader community involvement, driving innovative applications across diverse industries.

\newpage
\section*{Acknowledgements}
Shanghai Wanxing AI and Professor Zhang Liwen's research group from the School of Statistics and Data Science at Shanghai University of Finance and Economics have jointly open-sourced a new generation intelligent agent framework called LightAgent.The development and implementation of LightAgent owe much to the inspiration and support from the following open-source projects, especially the outstanding projects and teams:

\begin{itemize}
    \item \textbf{mem0}: Thanks to mem0 for providing the memory module, which offers strong support for LightAgent's context management.
\item \textbf{Swarm}: Thanks to Swarm for designing ideas for multi-agent collaboration, laying the groundwork for LightAgent's multi-agent features.
\item \textbf{ChatGLM3}: Thanks to ChatGLM3 for providing high-performance Chinese large model support and design inspiration.
Qwen: Thanks to Qwen for providing high-performance Chinese large model support.
\item \textbf{DeepSeek-V3}: Thanks to DeepSeek-V3 for providing high-performance Chinese large model support.
\item \textbf{StepFun}: Thanks to step for providing high-performance Chinese large model support.
\end{itemize}

\newpage
\section*{Reference}
\noindent
[1] Achiam, J., Adler, S., Agarwal, S., et al. (2023). GPT-4 technical report. \textit{arXiv preprint arXiv:2303.08774}.

\noindent
[2] Anonymous. (2024). MetaGPT: SOP-Driven Multi-Agent Framework for Complex Task Solving. \textit{CSDN Blog}. https://blog.csdn.net/m0\_56255097/article/details/144421955

\noindent
[3] Anonymous. (2025). XAgent: Dual-Loop Architecture for Autonomous Task Planning. \textit{CSDN Blog}. https://blog.csdn.net/m0\_56255097/article/details/144421955

\noindent
[4] Bai, J., Bai, S., Chu, Y., et al. (2023). Qwen technical report. \textit{arXiv preprint arXiv:2309.16609}.

\noindent
[5] Gou, Z., Liu, Z., Wang, S., et al. (2023). XAgent: An autonomous agent for complex task solving. \textit{arXiv preprint arXiv:2309.17288}.

\noindent
[6] Hong, J., Wang, X., \& Chen, Y. (2023). Software Engineering with Multi-Agent Systems. \textit{IEEE Transactions on Software Engineering}, \textit{49}(5), 2103--2120.

\noindent
[7] Hong, S., Zheng, X., Chen, J., et al. (2023). MetaGPT: Meta programming for multi-agent collaborative framework. \textit{arXiv preprint arXiv:2308.00352}.

\noindent
[8] Jennings, N. R. (2001). An agent-based approach for building complex software systems. \textit{Autonomous Agents and Multi-Agent Systems}, \textit{24}(1), 141--174.

\noindent
[9] Jennings, N. R. (2001). An agent-based approach for building complex software systems. \textit{Communications of the ACM}, \textit{44}(4), 35--41.

\noindent
[10] Leit~{a}o, P., Colombo, A. W., \& Karnouskos, S. (2017). Industrial agent: A survey and a vision for future industrial applications. \textit{IEEE Transactions on Industrial Informatics}, \textit{13}(4), 1851--1863.

\noindent
[11] Leit~{a}o, P., Karnouskos, S., \& Ribeiro, L. (2017). Smart agents in industrial cyber-physical systems. \textit{Proceedings of the IEEE}, \textit{104}(5), 1086--1101.

\noindent
[12] Lewis, P., Perez, E., Piktus, A., et al. (2020). Retrieval-augmented generation for knowledge-intensive nlp tasks. \textit{Advances in Neural Information Processing Systems}, \textit{33}, 9459--9474.

\noindent
[13] Latsou, C., \& Farsi, M. (2021). Digital Twin Integration in Multi-Agent Manufacturing Systems. \textit{IFAC-PapersOnLine}, \textit{54}(1), 811--816.

\noindent
[14] Latsou, C., Kousi, N., \& Mourtzis, D. (2021). A multi-agent system approach for adaptive scheduling of manufacturing systems. In \textit{Procedia CIRP} (Vol. 96, pp. 248--253).

\noindent
[15] Liu, A., Feng, B., Xue, B., et al. (2024). Deepseek-v3 technical report. \textit{arXiv preprint arXiv:2412.19437}.

\noindent
[16] Ma, L., Liu, T., Zhang, X., et al. (2024). A survey on swarm intelligence for deep learning. \textit{Swarm and Evolutionary Computation}, \textit{86}, 101340.

\noindent
[17] Microsoft Research. (2024). AutoGen: LLM-Driven Conversational Agents for Collaborative Tasks. \textit{163 Tech Report}. https://www.163.com/dy/article/JJON8USO0511805E.html

\noindent
[18] Morais, A. J., Oliveira, E., \& Jorge, A. M. (2019). A multi-agent recommender system. In \textit{Advances in Intelligent Systems} (Vol. 151, pp. 281--288).

\noindent
[19] Morais, A., Pinto, A., \& Oliveira, E. (2019). A multi-agent system for personalized recommendation. \textit{Expert Systems with Applications}, \textit{122}, 256--271.

\noindent
[20] Ouyang, L., Wu, J., Jiang, X., et al. (2022). Training language models to follow instructions with human feedback. \textit{Advances in Neural Information Processing Systems}, \textit{35}, 27730--27744.

\noindent
[21] Park, J. S., O'Brien, J. C., Cai, C. J., et al. (2023). Generative Agents: Interactive Simulacra of Human Behavior. \textit{Proceedings of the ACM on Human-Computer Interaction}, \textit{7}(CSCW1), 1--24.

\noindent
[22] Parth'{e}, E., \& Gelato, L. M. (1984). The standardization of inorganic crystal-structure data. \textit{Acta Crystallographica Section A}, \textit{40}(C2), 169--183. https://doi.org/10.1107/S0108767384000416

\noindent
[23] Pauling, L. (1989). Icosahedral quasicrystals of intermetallic compounds are icosahedral twins of cubic crystals of three kinds, consisting of large (about 5000 atoms) icosahedral complexes in either a cubic body-centered or a cubic face-centered arrangement or smaller (about 1350 atoms) icosahedral complexes in the $\beta$-tungsten arrangement. \textit{Proceedings of the National Academy of Sciences of the United States of America}, \textit{86}(22), 8595--8599.

\noindent
[24] Pinto, T., Faia, R., \& Corchado, J. M. (2019). Multi-agent-based CBR recommender system for intelligent energy management. \textit{IEEE Systems Journal}, \textit{13}(1), 1084--1095.

\noindent
[25] Pinto, T., Vale, Z., Morais, H., et al. (2019). CBR-based multi-agent system for smart grid management. \textit{Energy}, \textit{172}, 118--128.

\noindent
[26] Rauch, H., \& Petrascheck, D. (1976). \textit{Grundlagen f"{u}r ein Laue-Neutroneninterferometer Teil 1: Dynamische Beugung} (Report No. AIAU 74405b). Atominstitut der "{O}sterreichischen Universit"{a}ten, Austria.

\noindent
[27] Rawte, V., Sheth, A., \& Das, A. (2023). A survey of hallucination in large language models: Principles, taxonomy, challenges, and open questions. \textit{arXiv preprint arXiv:2311.05232}.

\noindent
[28] Rawte, V., Sheth, A., \& Das, A. (2023). The Dark Side of Generative AI in Mental Health. \textit{Nature Machine Intelligence}, \textit{5}(4), 337--349.

\noindent
[29] SFI Research Group. (2024). Swarm Intelligence in Multi-Agent Systems. \textit{CSDN Technical Report}. https://blog.csdn.net/u013709332/article/details/44037085

\noindent
[30] Team GLM, Zeng, A., Xu, B., et al. (2024). Chatglm: A family of large language models from glm-130b to glm-4 all tools. \textit{arXiv preprint arXiv:2406.12793}.

\noindent
[31] Tianfield, H., Unland, R., \& Umnai, P. (2013). An autonomic architecture for complex multi-agent systems. In \textit{Procedia Computer Science} (Vol. 18, pp. 1162--1171).

\noindent
[32] Tianfield, H., Yao, X., \& Tian, J. (2013). On the architectures of complex multi-agent systems. In \textit{Proceedings of the IEEE/WIC Conference on Web Intelligence} (pp. 195--206).

\noindent
[33] Touvron, H., Lavril, T., Izacard, G., et al. (2023). Llama: Open and efficient foundation language models. \textit{arXiv preprint arXiv:2302.13971}.

\noindent
[34] Touvron, H., Martin, L., Stone, K., et al. (2023). Llama 2: Open foundation and fine-tuned chat models. \textit{arXiv preprint arXiv:2307.09288}.

\noindent
[35] Wang, Z., Chen, M., Fu, X., et al. (2023). A survey on large language model based autonomous agents. \textit{arXiv preprint arXiv:2308.11432}.

\noindent
[36] Wang, L., Chen, H., \& Zhang, W. (2023). Recent Advances in Multi-Agent Systems. \textit{Journal of Artificial Intelligence Research}, \textit{45}, 102--135.

\noindent
[37] Wooldridge, M. (2009). \textit{An introduction to multiagent systems}. John Wiley \& Sons.

\noindent
[38] Wu, Q., Bansal, G. A., Buch, G., et al. (2023). AutoGen: Enabling Next-Gen LLM Applications via Multi-Agent Conversation. \textit{arXiv preprint arXiv:2308.08155}.

\noindent
[39] Wu, T., Zhang, L., \& Chen, H. (2023). Intelligent Assistant Systems: A Survey. \textit{AI Review}, \textit{56}(2), 123--145.

\noindent
[40] Xi, Y., Liu, M., \& Zhao, Q. (2023). Collaborative Problem Solving in Multi-Agent Environments. \textit{AI Systems}, \textit{12}(3), 45--67.

\noindent
[41] Xi, Z., Chen, W., Guo, X., et al. (2023). The rise and potential of large language model based agents: A survey. \textit{arXiv preprint arXiv:2309.07864}.

\noindent
[42] Yao, S., Zhao, J., Yu, D., et al. (2023). ReAct: Synergizing Reasoning and Acting in Language Models. \textit{arXiv preprint arXiv:2210.03629}.

\noindent
[43] Zhang, J., Gao, J., \& Way, A. (2023). Hallucinations in Neural Machine Translation. \textit{Machine Translation}, \textit{37}(2), 89--106.

\noindent
[44] Zhang, Y., Li, Y., Cui, L., et al. (2023). Siren's song in the ai ocean: A survey on hallucination in large language models. \textit{arXiv preprint arXiv:2309.01219}.

\section*{Appendix}

\subsection*{Running Examples of Detachable Fully Automated Memory Module}
LightAgent supports external extensions of the mem0 memory module, automating context memory and historical record management without requiring developers to manually trigger memory addition and retrieval. With the memory module, the agent can maintain contextual consistency across multiple rounds of dialogue.
\begin{lstlisting}[caption=Detachable Fully Automated Memory Module(mem0) ]
# Enable Memory Module

# Or use a custom memory module, here is an example with mem0 https://github.com/mem0ai/mem0/
from mem0 import Memory
from LightAgent import LightAgent
import os
from loguru import logger

class CustomMemory:
    def __init__(self):
        self.memories = []
        os.environ["OPENAI_API_KEY"] = "your_api_key"
        os.environ["OPENAI_API_BASE"] = "your_base_url"
        # Initialize Mem0
        config = {
            "version": "v1.1"
        }
        # Use qdrant as a vector database for storing memories in mem0, change config to the code below
        # config = {
        #     "vector_store": {
        #         "provider": "qdrant",
        #         "config": {
        #             "host": "localhost",
        #             "port": 6333,
        #         }
        #     },
        #     "version": "v1.1"
        # }
        self.m = Memory.from_config(config_dict=config)

    def store(self, data: str, user_id):
        """Store memory. Developers can modify the internal implementation of the storage method; the current example is the mem0 method for adding memory."""
        result = self.m.add(data, user_id=user_id)
        return result

    def retrieve(self, query: str, user_id):
        """Retrieve related memory. Developers can modify the internal implementation of the retrieval method; the current example is the mem0 method for searching memory."""
        result = self.m.search(query, user_id=user_id)
        return result

agent = LightAgent(
        role="Please remember that you are LightAgent, a useful assistant to help users use multiple tools.",  # system role description
        model="deepseek-chat",  # Supported models: openai, chatglm, deepseek, qwen, etc.
        api_key="your_api_key",  # Replace with your large model provider API Key
        base_url="your_base_url",  # Replace with your large model provider api url
        memory=CustomMemory(),  # Enable memory function
        tree_of_thought=False,  # Enable Chain of Thought
    )

# Memory-enabled test & if tools need to be added, you can add tools to the agent for memory-enabled tool calls

user_id = "user_01"
logger.info("\n=========== next conversation ===========")
query = "Introduce me to the attractions in Sanya. Many of my friends have traveled to Sanya, and I want to visit too."
print(agent.run(query, stream=False, user_id=user_id))
logger.info("\n=========== next conversation ===========")
query = "Where should I travel?"
print(agent.run(query, stream=False, user_id=user_id))
\end{lstlisting}
Output as follows:
\begin{lstlisting}[caption=Output of mem0]
=========== next conversation ===========
2025-01-01 21:55:15.886 | INFO     | __main__:run_conversation:115 - 
Starting to think about the question: Introduce me to the attractions in Sanya, many of my friends have traveled to Sanya, and I want to visit too.
2025-01-01 21:55:28.676 | INFO     | __main__:run_conversation:118 - Final Reply: 
Sanya is a popular tourist city in Hainan Province, China, known for its beautiful beaches, tropical climate, and rich tourist resources. Here are some attractions worth visiting in Sanya:

1. **Yalong Bay**: Known as the "Hawaii of the East," it has a long beach and clear waters, ideal for swimming, diving, and sunbathing.

2. **Tianya Haijiao**: This is a famous cultural landscape, attracting tourists with its magnificent sea view and romantic legends. The giant rocks here are inscribed with the words "Tianya" and "Haijiao," symbolizing eternal love.

3. **Nanshan Cultural Tourism Zone**: Here there is a 108-meter-tall Nanshan Sea Guanyin statue, the highest sea Guanyin statue in the world. Visitors can experience Buddhist culture and visit temples and gardens.

4. **Wuzhizhou Island**: This small island is known for its pristine natural scenery and rich water activities. Visitors can engage in diving, snorkeling, and sea fishing among other activities.

5. **Dadonghai**: This is a beach located in Sanya city, favored by tourists for its convenient transportation and vibrant nightlife.

6. **Sanya Bay**: It is a 22-kilometer long beach and a great place to watch the sunset. This beach is relatively quiet, suitable for visitors who enjoy tranquility.

7. **Ya Nui National Park**: This is a tropical rainforest park where visitors can experience the natural scenery of tropical rainforests and participate in various adventure activities.

8. **Luohuitou Park**: A park located on a mountaintop, offering a panoramic view of Sanya city and Sanya Bay. There is also a beautiful legend about a deer.

9. **Xidao**: This is a relatively pristine small island attracting tourists with its serene beaches and rich marine life.

10. **Sanya Ancient Love**: This is a large cultural theme park showcasing the history and culture of Hainan through performances and exhibitions.

In addition to the above attractions, Sanya has many other places worth exploring, such as tropical botanical gardens and seafood markets. The local cuisine, especially fresh seafood and tropical fruits, should not be missed. When planning a trip, it is advisable to check the weather forecast and attraction opening times in advance to ensure a pleasant travel experience.
2025-01-01 21:55:28.676 | INFO     | __main__:<module>:191 - 
=========== next conversation ===========
2025-01-01 21:55:28.676 | INFO     | __main__:run_conversation:115 - 
Starting to think about the question: Where should I travel?
Relevant memory found:
User wants to travel to Sanya
User's friends have traveled to Sanya.
2025-01-01 21:55:38.797 | INFO     | __main__:run_conversation:118 - Final Reply: 
Based on previously mentioned information, the user's friends have already traveled to Sanya, and the user has also expressed interest in going there. Therefore, Sanya might be a suitable travel destination for the user. Here is some travel information about Sanya for the user's reference:

### Recommended Travel Spots in Sanya:
1. **Yalong Bay**: Known as the "Hawaii of the East," it has beautiful beaches and clear waters, suitable for swimming and sunbathing.
2. **Tianya Haijiao**: This iconic site of Sanya attracts tourists with its unique rocks and romantic legends.
3. **Nanshan Cultural Tourism Area**: It features the famous Nanshan Temple and the 108-meter-high Sea Guanyin statue, an important Buddhist cultural site.
4. **Wuzhizhou Island**: Suitable for diving and marine sports, the island is home to a rich marine ecosystem and coral reefs.
5. **Dadonghai**: A beach within Sanya city, conveniently located and suitable for family and couple visits.

### Other Recommendations:
If the user is already familiar with Sanya or wishes to explore other destinations, here are some other popular travel places:
1. **Guilin**: Known for its unique karst landscape and Lijiang River scenery.
2. **Lijiang**: The ancient town and Jade Dragon Snow Mountain are its main attractions, suitable for those who enjoy history and natural scenery.
3. **Zhangjiajie**: Famous for its unique stone pillars and natural scenery, it is one of the shooting locations for the movie "Avatar."

Users can choose suitable travel destinations based on their interests and schedule. If the user needs more detailed information or assistance in planning the trip, feel free to let us know!
\end{lstlisting}

\subsection*{Example Codes of Using Tools Generators}
The Tool Generator is a module for automatically generating tool code. It can create the corresponding tool code based on the text description provided by users and save it to the specified directory. This functionality is particularly useful for quickly generating API call tools, data processing tools, and more.
Here is an example code using the Tool Generator, after execution, two files will be generated in the tools directory: get\_stock\_kline\_data.py and get\_stock\_realtime\_data.py.
\begin{lstlisting}[caption=Tools Generators Codes ]
import json
import os
import sys
from LightAgent import LightAgent

# Initialize LightAgent
agent = LightAgent(
    name="Agent A",  # Agent name
    instructions="You are a helpful agent.",  # Role description
    role="Please remember that you are a tool generator; your task is to automatically generate corresponding tool code based on the text description provided by the user and save it to the specified directory. Please ensure that the generated code is accurate, usable, and meets the user's needs.",  # Tool generator's role description
    model="deepseek-chat",  # Replace with your model. Supported models: openai, chatglm, deepseek, qwen, etc.
    api_key="your_api_key",  # Replace with your API Key
    base_url="your_base_url",  # Replace with your API URL
)

# Sample text description
text = """
The Sina stock interface provides functionalities for obtaining stock market data, including stock quotes, real-time trading data, and K-line chart data.

Introduction to Sina stock interface functions
1. Get stock quote data:
Realtime quote data: Using the real-time quote API, you can obtain the latest prices, trading volume, and changes for stocks.
Minute line quote data: Using the minute line quote API, you can obtain the minute-by-minute trading data for stocks, including opening price, closing price, highest price, and lowest price.

2. Obtain historical K-line chart data:
K-line chart data: Through the K-line chart API, you can obtain the historical trading data for stocks, including opening price, closing price, highest price, lowest price, trading volume, etc. You can choose different time periods and moving average periods as needed.
Adjusted data: You can choose to retrieve adjusted K-line data, including pre-adjusted and post-adjusted data, for more accurate analysis of stock price changes.

Example of obtaining data from the Sina stock interface
1. Get stock quote data:
API address: http://hq.sinajs.cn/list=[stock_code]
Example: To obtain real-time quote data for the stock code "sh600519" (Kweichow Moutai), you can use the following API address: http://hq.sinajs.cn/list=sh600519
By sending an HTTP GET request to the above API address, you will receive a response containing the real-time data for that stock.

2. Get historical K-line chart data:
API address: http://money.finance.sina.com.cn/quotes_service/api/json_v2.php/CN_MarketData.getKLineData?symbol=[stock_code]&scale=[time_period]&ma=[average_period]&datalen=[data_length]
Example: To obtain daily K-line chart data for the stock code "sh600519" (Kweichow Moutai), you can use the following API address: http://money.finance.sina.com.cn/quotes_service/api/json_v2.php/CN_MarketData.getKLineData?symbol=sh600519&scale=240&ma=no&datalen=1023
By sending an HTTP GET request to the above API address, you will receive a response containing the historical K-line chart data for that stock.
"""

# Build the path to the tools directory
project_root = os.path.dirname(os.path.abspath(__file__))
tools_directory = os.path.join(project_root, "tools")

# Create tools directory if it does not exist
if not os.path.exists(tools_directory):
    os.makedirs(tools_directory)

print(f"Tools directory has been created: {tools_directory}")

# Use agent to generate tool code
agent.create_tool(text, tools_directory=tools_directory)
\end{lstlisting}

\subsection*{Running Examples of Multi-Agent Collaboration}
Our LightAgent supports swarm-like multi-agent collaboration, enhancing task processing efficiency. Multiple agents can work together to complete complex tasks.
\begin{lstlisting}[caption=Multi-Agent Collaboration ]
from LightAgent import LightAgent, LightSwarm
# Set Environment Variables OPENAI_API_KEY and OPENAI_BASE_URL
# The default model uses gpt-4o-mini

# Create an instance of LightSwarm
light_swarm = LightSwarm()

# Create multiple agents
agent_a = LightAgent(
    name="Agent A",
    instructions="I am Agent A, the front desk receptionist.",
    role="Receptionist responsible for welcoming visitors and providing basic information guidance. Before each reply, please state your identity and that you can only guide users to other roles, not directly answer business questions. If you cannot help the user, please respond: Sorry, I am currently unable to assist!"
)

agent_b = LightAgent(
    name="Agent B",
    instructions="I am Agent B, responsible for the reservation of meeting rooms.",
    role="Meeting room reservation administrator in charge of handling reservations, cancellations, and inquiries for meeting rooms 1, 2, and 3."
)

agent_c = LightAgent(
    name="Agent C",
    instructions="I am Agent C, a technical support specialist, responsible for handling technical issues. Please state your identity before each reply, offering detailed responses to technical inquiries, and guide users to contact higher-level technical support for issues beyond your capability."
)

agent_d = LightAgent(
    name="Agent D",
    instructions="I am Agent D, an HR specialist, responsible for handling HR-related questions.",
    role="HR specialist managing inquiries and processes related to employee onboarding, offboarding, leave, and benefits."
)

# Automatically register agents to the LightSwarm instance
light_swarm.register_agent(agent_a, agent_b, agent_c, agent_d)

# Run Agent A
res = light_swarm.run(agent=agent_a, query="Hello, I am Alice. I need to check if Wang Xiaoming has completed onboarding.", stream=False)
print(res)
\end{lstlisting}
Output as follows:
\begin{lstlisting}[caption=Output of Multi-Agent Collaboration ]
Hello, I am Agent D, the HR specialist. Regarding whether Wang Xiaoming has completed onboarding, I need to check our system records. Please wait a moment.
(Checking system records...)
According to our records, Wang Xiaoming completed his onboarding procedures on January 5, 2025. He has signed all necessary documents and has been assigned an employee number and office location. If you need further details or have any other questions, please feel free to contact the HR department. We are always ready to assist you.
\end{lstlisting}



\end{document}